\documentclass[cameraready]{Interspeech}
\usepackage{amsmath,graphicx,hyperref}
\usepackage{booktabs} 
\usepackage{tabularx}  
\usepackage{multirow}  
\usepackage{caption}
\usepackage{subcaption}
\usepackage{microtype}
\usepackage{tcolorbox}
\usepackage{cite}
\hypersetup{
    colorlinks=true,
    citecolor=blue,
    linkcolor=red,
    urlcolor=red
}

\title{Speech-based Psychological Crisis Assessment using LLMs}

\author[affiliation={1}, orcid=0009-0003-2174-0729, equalcontribution]{Terumi}{Chiba}
\author[affiliation={2,3}, equalcontribution]{Yang}{Luo}
\author[affiliation={1}, orcid=0009-0009-0485-6141]{Ziyun}{Cui}
\author[affiliation={2,3}, orcid=0000-0002-1544-040X, correspondingauthor]{Yongsheng}{Tong}
\author[affiliation={1}, orcid=0000-0002-7730-5131, correspondingauthor]{Chao}{Zhang}

\address{
  $^1$ Department of Electronic Engineering, Tsinghua University, Beijing, China \\
  $^2$Peking University Huilongguan Clinical Medical School, Beijing, China \\
  $^3$WHO Collaborating Centre for Research and Training in Suicide Prevention
}

\email{qyxh25@mails.tsinghua.edu.cn, cz277@tsinghua.edu.cn}


\usepackage{comment}

\begin{document}

\maketitle

\begin{abstract}
Psychological support hotlines provide critical support for individuals experiencing mental health emergencies, yet crisis-level assessment largely relies on human operators whose judgments may vary with training and staffing constraints. This paper studies automated three-way crisis-level classification from authentic hotline speech under privacy-sensitive and data-scarce conditions. We propose an LLM-based pipeline that converts non-verbal acoustic cues into explicit textual evidence through paralinguistic injection, enabling a text LLM to use vocal affect that is otherwise lost in ASR transcripts. We also explore reasoning-enhanced training, where LLM-generated diagnostic rationales are used as auxiliary supervision to regularise the classification objective. With chunk-based data augmentation and 5-fold cross-validation, our system achieves a macro F1-score of 0.802 and an accuracy of 0.805, outperforming acoustic, zero-shot LLM, and speech-aware LLM baselines.
\end{abstract}

\section{Introduction}

Psychological crises are acute states in which perceived difficulties exceed coping capacity and may escalate to self-harm, suicide, or harm to others~\cite{CrisisStratChap1}. Psychological support hotlines play a key role in mitigating such emergencies~\cite{zabelski2023crisis}, but triage is constrained by limited staff resources and subjective severity ratings across operators with different training and experience~\cite{tyson2016preventing, wang2020hotline, hoffberg2020effectiveness, you2021study}. These challenges motivate automated crisis-level assessment as decision support for more consistent triage and resource planning.

Prior mental-health assessment studies on hotline or counselling data have used handcrafted acoustic features~\cite{su2025acoustic}, Transformer-based text classifiers~\cite{broadbent2023machine, imel2024machine, song2024exploratory}, and more recently LLMs for related tasks such as Alzheimer’s disease diagnosis, depression detection, and suicide risk prediction~\cite{feng2023large, ikeuchi2025efficient, qorich2024advanced, chen2024deep}. Most hotline-based studies focus on suicide risk assessment, while Deng et al.\cite{deng2025evaluating} explore binary psychological crisis classification from real-world crisis-call transcriptions. However, this text-only formulation overlooks paralinguistic information in speech, which is important in clinical practice\cite{CrisisStratChap3}.

In this paper, we study three-way psychological crisis level classification from authentic hotline speech under privacy-sensitive and data-scarce conditions. Our experiments are designed to examine three questions: (i) whether explicit textualisation of paralinguistic cues provides useful evidence beyond plain ASR transcripts; (ii) whether a text-based LLM with injected paralinguistic evidence is more effective than directly fine-tuning a SpeechLLM in this data-scarce setting; and (iii) whether LLM-generated diagnostic rationales can serve as auxiliary supervision for crisis-level classification. To this end, we propose an LLM-based framework that leverages both linguistic and paralinguistic information in speech. Specifically, a SpeechLLM is used to extract non-verbal paralinguistic cues, which are then inserted into ASR-generated transcripts through \textit{paralinguistic injection}, enabling the downstream text LLM to condition on vocal cues that would otherwise be lost in standard transcripts. We further explore \textit{reasoning-enhanced training}, where LLM-generated diagnostic reasoning chains~\cite{teng2025enhancing, patil2025cognitive} are used as auxiliary training targets to regularise the classifier. Finally, we employ chunk-based data augmentation to mitigate the scarcity of authentic clinical data. With call-level 5-fold cross-validation, the proposed framework achieves a macro F1-score of 0.802 and an accuracy of 0.805, outperforming acoustic, zero-shot LLM, and SpeechLLM baselines. We view this system as decision support for more consistent triage and resource planning under human oversight, rather than as a replacement for clinical assessment.

The remainder of this paper is structured as follows: Section~\ref{sec:dataset} describes the dataset and ethical considerations, Section~\ref{sec:methods} details the proposed framework, Sections~\ref{sec:exp_setup} and \ref{sec:results} present the experimental setup and results, and we conclude in Sec.~\ref{sec:conclusion}.

\section{Task and Dataset}
\label{sec:dataset}

We collected the dataset from a Chinese psychological support hotline, consisting of 154 calls, where all calls were annotated by two experts, both the Chief Director of Hotline Assessment, into three classes: (\texttt{0}) no crisis, (\texttt{1}) low crisis, and (\texttt{2}) medium-to-high crisis, and inter-annotator agreement was assessed to verify annotation consistency. The callers span diverse demographics (gender, age, occupation) and discuss a wide range of personal issues, including depression, anxiety, relationship stress, and suicidal ideation, \textit{etc.} The spontaneous and emotionally rich nature of these recordings makes the dataset particularly valuable for crisis detection research. 

\begin{figure}[htbp]
    \centering

    \begin{minipage}[b]{0.49\columnwidth}
        \centering
        \subcaptionbox{Age Distribution\label{fig:age}}
        {\includegraphics[width=\textwidth]{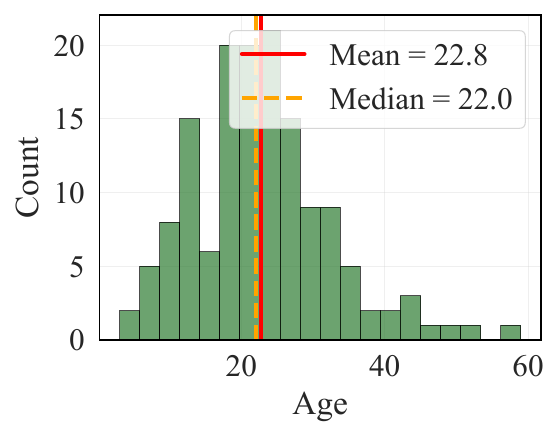}}
    \end{minipage}
    \hfill
    \begin{minipage}[b]{0.49\columnwidth}
        \centering
        \subcaptionbox{Duration Distribution\label{fig:duration}}
        {\includegraphics[width=\textwidth]{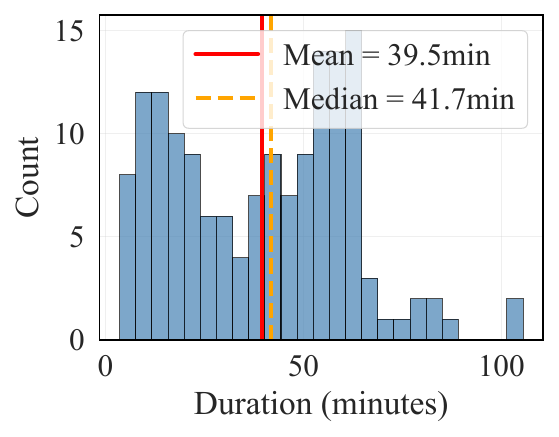}}
    \end{minipage}
    \vspace{2mm} 

    \begin{minipage}[b]{0.45\columnwidth}
        \centering
        \subcaptionbox{Crisis Level Distribution\label{fig:crisis}}
        {\includegraphics[width=\textwidth]{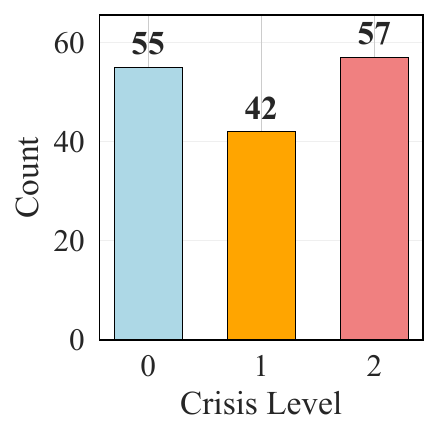}}
    \end{minipage}
    \hfill
    \begin{minipage}[b]{0.45\columnwidth}
        \centering
        \subcaptionbox{Duration vs. Crisis Level\label{fig:box}}
        {\includegraphics[width=\textwidth]{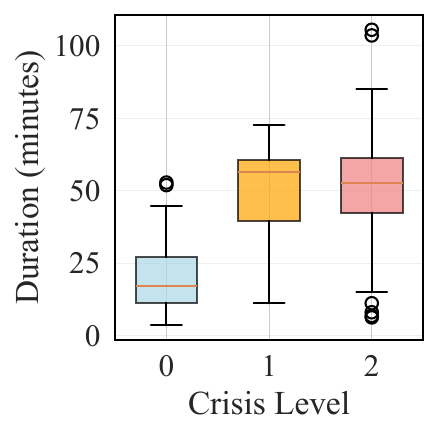}}
    \end{minipage}
    \caption{Data statistics: (a) age distribution, (b) duration distribution, (c) crisis level distribution, and (d) relation between duration and crisis level.}
    \label{fig:data_stats}
\end{figure}

Fig.~\ref{fig:data_stats} provides a statistical overview of the dataset. The total length of dataset is approximately 100 hours, with 39.49 $\pm$ 22.3 minutes per sample on average. The distribution of call durations is bimodal, indicating a prevalence of both short and long conversations. The age distribution approximates a normal curve centered around 22, although age information is missing for eight callers. Among the 154 instances, gender metadata is available for 99 callers, comprising 57 males and 42 females. The three crisis levels -- ``no crisis'' (55 calls), ``low crisis'' (42 calls), and ``medium-to-high crisis'' (57 calls) -- are relatively balanced. Calls with crisis tend to be longer than non-crisis calls, shown in the Fig.~\ref{fig:box}.

All audio and transcripts were processed under the hotline center's ethical guidelines to protect personal information.

\section{Methods}
\label{sec:methods}


\subsection{Overall Framework}
\label{sec:framework}
\begin{figure}[htbp]
    \centering
    \includegraphics[width=1\columnwidth]{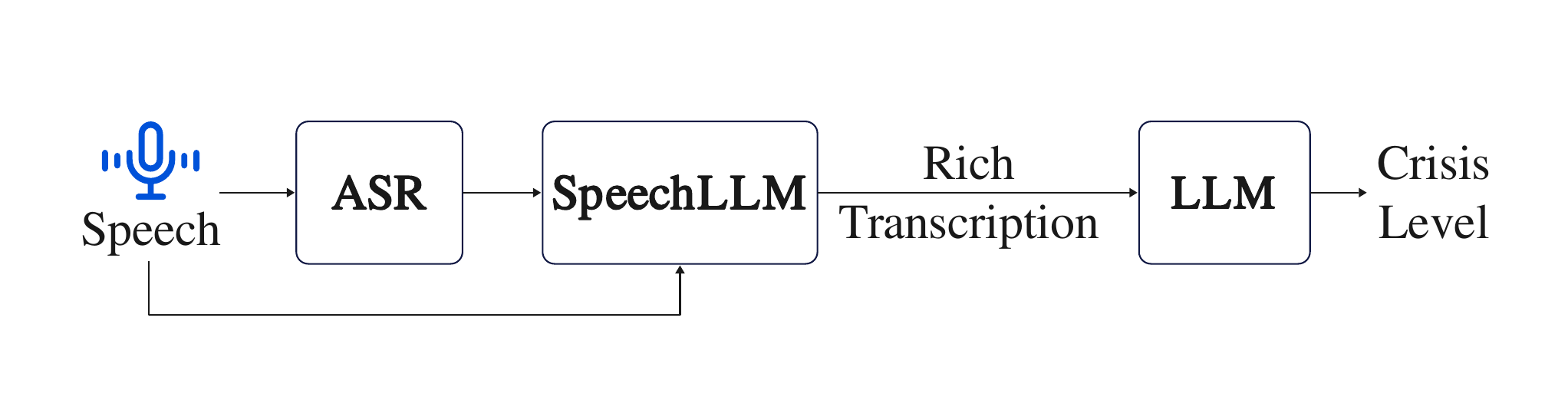} 
    \caption{The overall framework of the proposed method. A raw hotline recording is converted into a transcript by ASR and into textual paralinguistic cues by SpeechLLM. The two outputs are combined into a paralinguistically enriched transcript, which is then used by an LLM to predict the crisis level.}
    \label{fig:model_framework} 
\end{figure}

Our proposed crisis classification framework processes conversational hotline data through a two-step pipeline comprising multimodal data preprocessing and an LLM-based classifier. 
As illustrated in Fig.~\ref{fig:model_framework}, the raw audio is first transcribed into text using an ASR system (Paraformer-zh\footnote{\url{https://huggingface.co/funasr/paraformer-zh}}~\cite{gao2022paraformer}), and injected paralinguistic descriptions through SpeechLLM (Step-Audio-R1\footnote{\url{https://huggingface.co/stepfun-ai/Step-Audio-R1.1}}~\cite{tian2025step}), resulting in a paralinguistically enriched transcription. The resulting \textit{rich transcription} was then fed into a fine-tuned LLM to predict the crisis level.

\subsection{Paralinguistic Injection}
Our paralinguistic injection is guided by the triage assessment form (TAF) for crisis intervention~\cite{CrisisStratChap3}, which evaluates crisis severity across affective, behavioural, and cognitive domains, providing explicit impairment guidelines and categorising primary, secondary, and tertiary crisis-related emotions. 

Given that ASR-derived transcriptions omit non-verbal cues, we inject paralinguistic annotations into the text for subsequent LLM processing. Step-Audio-R1~\cite{tian2025step} is applied in this module.
Guided by TAF's affective domain constraints, we prompt the speech model to focus on identifying and extracting the core crisis-related emotions and their corresponding acoustic manifestations (\textit{e.g.}, sobbing, sighing, and trembling voice). 

An example of a raw ASR transcription and its paralinguistically enriched transcription is shown below\footnote{The original transcripts and model outputs in our dataset are in Mandarin Chinese; the examples presented in this paper have been translated into English for illustration.}:

\begin{tcolorbox}[colback=gray!5, colframe=gray!50!black, title=\textbf{Example: Paralinguistic Injection}]
\textit{Raw ASR Transcription:} ``I was forced to move out of the school dorm...''\\
\textit{Paralinguistically Enriched Transcription:} ``I was forced to move out of the school dorm... [Trembling voice, obvious sobbing, low volume, expressing extreme sadness and helplessness; Emotion: Sadness $>$ Anxiety]''

\end{tcolorbox}

In this example, the output text effectively reduces the ambiguity of purely lexical interpretation (\textit{e.g.}, “move out” could be neutral in text) by revealing the underlying distress intensity. The paralinguistically enriched transcript makes the speaker's affect explicit and aligned with the corresponding utterance, allowing the downstream LLM to ground its crisis reasoning in both semantic content and clinically salient vocal evidence.

\subsection{Reasoning Enhanced Training}

\begin{figure}[htbp]
    \centering
    \includegraphics[width=0.9\columnwidth]{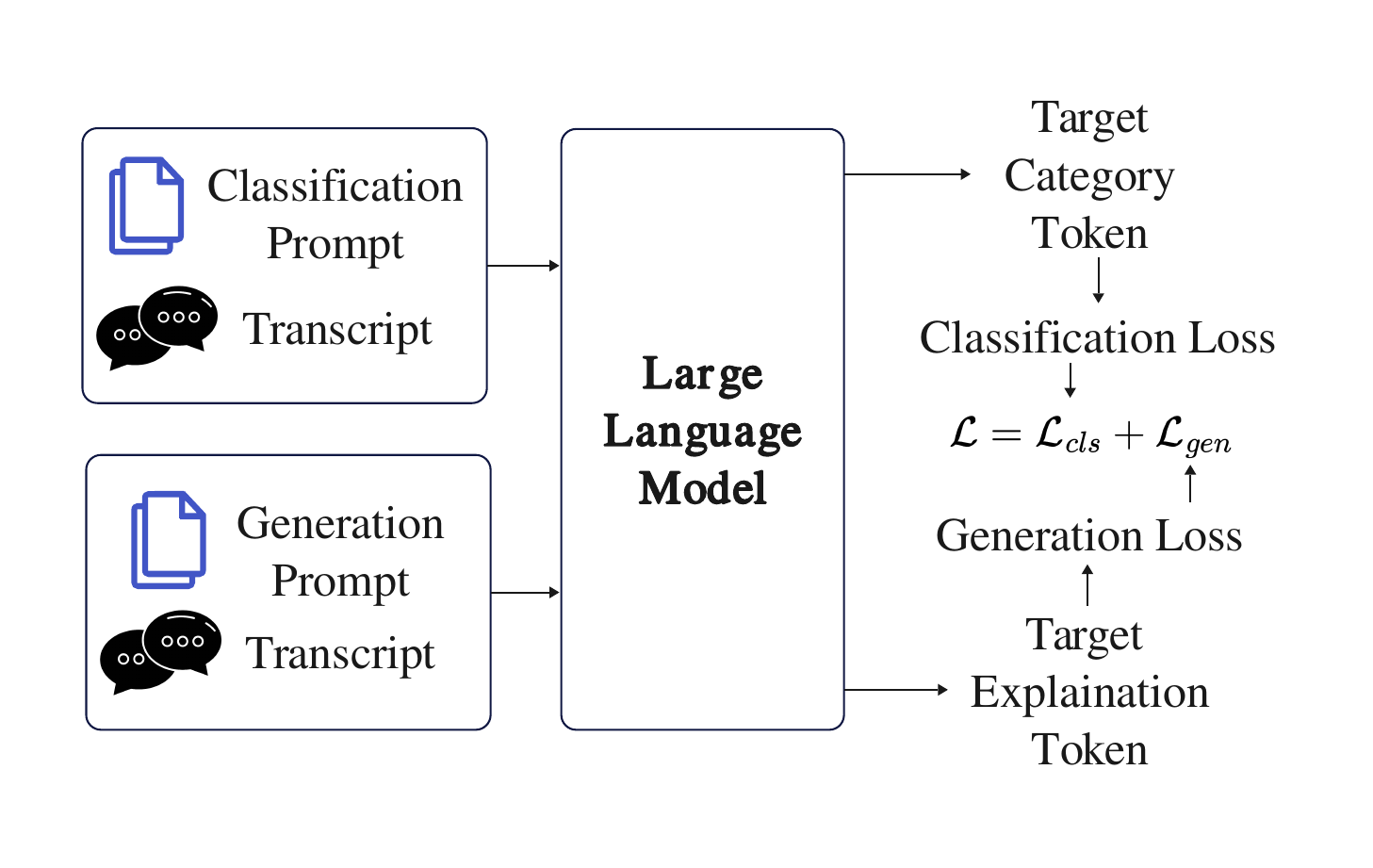} 
    \caption{Model training pipeline. The enriched transcript is used in two parallel forward passes: one with a classification prompt to predict the crisis level, and the other with a generation prompt to produce a TAF-constrained label explanation. The classification loss $\mathcal{L}_\text{cls}$ and generation loss $\mathcal{L}_\text{gen}$ are combined as the final training objective.}
    \label{fig:aux} 
\end{figure}

The model training pipeline is shown in Fig.~\ref{fig:aux}. The crisis classification was formulated as a next-token prediction problem. Specifically,  the paralinguistic-injected transcription is concatenated with a classification prompt (``Based on the emotional representations and the psychological support hotline dialogue provided above, I assess the caller’s crisis level (\texttt{0}, \texttt{1}, or \texttt{2}) as:''). The output probabilities of the three category tokens {`0', `1', `2'} were then extracted from the model's next-token distribution for computing cross-entropy loss against the classification label. The cross-entropy loss is applied to update the model parameters.

Building upon this classification objective, we further propose a \textit{reasoning enhanced training} strategy that introduces diagnostic reasoning generation as an auxiliary task, providing an auxiliary learning signal that encourages the model to attend to evidence aligned with TAF criteria.

It is important to note that the dataset contains expert annotations only for the final three-way crisis level, rather than separate expert labels for the Affective, Behavioural, and Cognitive impairment domains. The auxiliary reasoning targets are therefore not treated as clinical ground truth. Instead, they are LLM-generated, TAF-conditioned rationales used only as an auxiliary training signal. Our evaluation remains based solely on the expert-annotated crisis-level labels.

The target reasoning text is constructed according to the TAF’s scoring logic: we prompt the gpt-oss-120b model\footnote{\url{https://huggingface.co/openai/gpt-oss-120b}}~\cite{agarwal2025gpt} to generate step-by-step clinical rationales conditioned on (i) the paralinguistic enriched transcript, (ii) the ground-truth crisis label, and (iii) the explicit TAF criteria across the Affective, Behavioral, and Cognitive domains. This provides a structured auxiliary target intended to expose the model to TAF-aligned evidence patterns. An example of the target text is as follows:

\begin{tcolorbox}[colback=gray!5, colframe=gray!50!black, title=\textbf{Example: Diagnostic Reasoning}]
``The caller exhibits sadness and anxiety (e.g., sobbing tone, low pitch), but can still answer questions smoothly with limited emotional fluctuation, which corresponds to mild affective impairment (2/3), yielding an Affective score of 2. Although there is a history of self-harm and occasional impulsivity, the current behavior does not significantly inhibit daily functioning; the caller is still able to attend school and interact with others, matching mild behavioral impairment (2/3), resulting in a Behavioral score of 2. The caller can clearly describe emotions, causes, and future plans. Although thinking is influenced by negative emotions, there is no cognitive confusion or loss of decision-making control, representing mild cognitive impairment (2/3), leading to a Cognitive score of 2. The three-dimensional scores are 2/2/2, with a total score of 6. Therefore, the caller's crisis level is 0 (high autonomy, requires support only).''
\end{tcolorbox}

In practice, for each training sample we perform two parallel forward passes. The classification prompt was applied to get a classification loss $\mathcal{L}_\text{cls}$. The generation prompt was applied for a generation loss $\mathcal{L}_\text{gen}$, via teacher forcing on the target reasoning tokens. The final training objective is the sum of the two losses:
\begin{equation}
\mathcal{L} = \mathcal{L}_\text{cls} + \mathcal{L}_\text{gen}.
\end{equation}

\subsection{Data Augmentation}
Due to the limited number of our dataset (154 recordings), we employed data augmentation to improve model robustness and prevent overfitting. Each original call was divided into non-overlapping, contiguous 5-minute chunks, and the ASR transcripts were aligned with the same chunk boundaries. This preserves the complete temporal continuity and conversational flow within each chunk, thus providing the model with rich sequences without disrupting clinical context.
After augmentation, we obtained 900+ speech chunks in total.

At test time, all chunks from the same call were kept in the same cross-validation fold, and chunk-level predictions were aggregated by majority voting to obtain the final call-level prediction.

\section{Experimental Setup}
\label{sec:exp_setup}

\subsection{Evaluation Protocol}
A 5-fold cross-validation scheme was applied to ensure a robust evaluation. In each fold, the data was partitioned into an 80\% training set and a 20\% test set. We report the mean and standard deviation of the macro F1-score across all five folds as our primary performance metric. Accuracy is also reported.

\subsection{Baselines}
We established three baselines for comparison: a classical acoustic machine learning model, a zero-shot LLM with Enriched Transcript pipeline, and a SpeechLLM-based fine-tuning baseline.

\textbf{OpenSMILE Baseline:} This baseline represents traditional speech-based crisis detection using handcrafted paralinguistic descriptors. We extracted functional-level acoustic features using the eGeMAPSv02 feature set~\cite{eyben2015geneva} via OpenSMILE toolkit~\cite{eyben2010opensmile}. To eliminate confounding factors, we exclusively used the caller's audio, thoroughly isolating and removing the operator's voice. For classification, we trained a support vector machine configured with a radial basis function kernel, regularisation parameter $C=1.0$, $\gamma=\text{`scale'}$, and $\text{class\_weight}=\text{`balanced'}$.

\textbf{Zero-shot LLM with Enriched Transcript Baseline:} To evaluate how far a general-purpose LLM can go without any domain fine-tuning, we fed the paralinguistic enriched transcript into the gpt-oss-120b model, operating in its high-reasoning mode (\texttt{think=high}). The prompt included a detailed description of the TAF, instructing the model to estimate severity scores for the Affective (A), Behavioural (B), and Cognitive (C) domains. The final crisis level was ultimately determined by aggregating these three scores.

\textbf{SpeechLLM Baseline:} This baseline serves as an end-to-end speech-aware LLM that directly consumes audio, to test whether explicitly injecting paralinguistic cues into text provides advantages over modelling speech paralinguistics in the audio domain. Qwen2.5-Omni-7B\footnote{\url{https://huggingface.co/Qwen/Qwen2.5-Omni-7B}}~\cite{xu2025qwen2} model was applied as the backbone, and followed the same training strategy and data augmentation. The same 5-fold cross-validation was conducted, with the test set evaluated with the majority voting result.

\subsection{Implementation Details}

The system was implemented in PyTorch. We fine-tuned the Qwen2.5-7B-Instruct\footnote{\url{https://huggingface.co/Qwen/Qwen2.5-7B-Instruct}}~\cite{yang2024qwen2} model using parameter-efficient fine-tuning via low-rank adaptation (LoRA)~\cite{hu2022lora}. All experiments were conducted on a single NVIDIA A800 GPU. Optimisation was performed with AdamW (default settings), using a learning rate of $3 \times 10^{-5}$ under a cosine annealing schedule with a linear warmup over the first 10\% of training steps. We used a minibatch size of 1 with gradient accumulation over 16 steps (effective batch size 16). For LoRA, we set the rank $r=8$ and scaling factor $\alpha=64$, applying it to the attention linear projection layers, including the query, key, and value matrices and the output projections. All runs were trained until convergence with a fixed random seed to ensure reproducibility.

\section{Results and Discussion}
\label{sec:results}
In this section, the empirical results of our proposed crisis classification system alongside detailed discussions are presented. We evaluate the effectiveness of our proposed method and conduct ablation studies on key components.

\subsection{Main Result}
We first evaluate the overall effectiveness of the proposed framework. The results are presented in Table~\ref{tab:aug_analysis}, with our proposed system yields an accuracy of 0.805 and a macro F1-score of 0.802.
The zero-shot LLM pipeline resulted in a macro F1-score of 0.371, indicating that prompt-only reasoning is insufficient for reliable crisis assessment on authentic hotline data and that task-specific training with structured supervision is crucial. The OpenSMILE acoustic baseline got a macro F1-score of 0.471, demonstrating that handcrafted paralinguistic descriptors alone cannot capture the cross-contextual semantics required for nuanced crisis-level discrimination.

Notably, our method also surpasses the SpeechLLM fine-tuning baseline by an absolute gain of 0.251, despite both being trained under comparable data protocols and evaluation settings. This result aligns with recent findings that current SpeechLLMs still exhibit clear weaknesses in contextual paralinguistic reasoning and affect-grounded understanding~\cite{wang2025benchmarking, wang2025incorporating}. In contrast, our paralinguistic injection converts sequence-aligned acoustic evidence into explicit textual markers that a strong text LLM can reliably use during reasoning, avoiding the common failure mode where SpeechLLMs underutilise prosody unless specifically trained with prosody-aware supervision. Overall, the improvements over the baseline systems suggest that representing paralinguistic evidence in an explicit textual form can be effective for this data-scarce crisis-classification setting.

\begin{table}[t]
\centering
\caption{Performance comparisons. Accuracy and Macro F1 (mean $\pm$ std) reported in percentage across 5 folds.}
\label{tab:aug_analysis}
\begin{tabular}{lcc}
\toprule
\textbf{Configuration} & \textbf{Accuracy} & \textbf{Macro F1} \\
\midrule
Zero-shot LLM & & \\
\quad w/ Enriched Transcript & 0.455 & 0.371 \\
OpenSMILE & 0.486 $\pm$ 0.053  & 0.471 $\pm$ 0.062 \\
SpeechLLM & 0.564 $\pm$ 0.075 & 0.551 $\pm$ 0.079 \\
Ours  & \textbf{0.805} $\pm$ \textbf{0.061} & \textbf{0.802} $\pm$ \textbf{0.062} \\
\bottomrule
\end{tabular}
\end{table}

\subsection{Ablation}
Ablation studies were conducted to investigate the contributions of data augmentation, paralinguistic injection, and the auxiliary reasoning loss, with the results summarised in Table~\ref{tab:aux_loss}.

\textbf{Effect of Data Augmentation:} The most substantial performance degradation occurred when removing data augmentation, resulting in a 10.0\% absolute drop in the macro F1-score. This suggests that the model relies heavily on the augmented data to learn generalizable crisis patterns. By splitting calls into continuous, fixed-length chunks, the model is exposed to a larger pseudo-dataset and converges more stably. 
Although training on full, uncut sequences theoretically preserves context, the limited number of training samples can cause the model to overfit speaker-specific characteristics rather than learning generalisable crisis representations~\cite{panayotov2015librispeech}.

\textbf{Effect of Paralinguistic Injection:} Omitting the paralinguistic injection module led to a 4.1\% decrease in the macro F1-score. This suggests that plain ASR transcripts may omit emotionally salient information needed for fine-grained crisis assessment. Explicitly translating these vital non-verbal cues into text effectively bridges the modality gap, providing the purely text-based LLM with essential Affective domain information.

Additionally, we conducted an additional experiment to further demonstrate the effectiveness of paralinguistic injection. Specifically, we extracted emotion embeddings for each utterance using emotion2vec\footnote{\url{https://huggingface.co/emotion2vec/emotion2vec_plus_large}}~\cite{ma2024emotion2vec}. 

The resulting emotion embeddings were first passed through an MLP adapter to project from the emotion representation space into the LLM embedding space, and then concatenated with the raw ASR text embeddings along the sequence-length dimension before being fed into Transformer blocks for joint processing.
With all other experimental settings kept unchanged, replacing the original paralinguistic injection with emotion2vec-based adaptation resulted in a similar performance degradation, yielding a 4.2\% decrease in macro-F1 compared to the full model. 
This may stem from the dataset being too small to reliably train an adapter from scratch, further highlighting the effectiveness of our explicit injection strategy.

\textbf{Effect of Auxiliary Reasoning Loss:} Finally, removing the auxiliary reasoning loss resulted in a modest performance reduction of 1.7\% in the macro F1-score. This supports our hypothesis that generating explicit clinical reasoning serves as a beneficial soft constraint. By training the model to generate TAF-aligned rationales as an auxiliary objective rather than jumping directly to a label, the auxiliary loss improves classification accuracy by modelling the latent reasoning that links speech content in hotline dialogues to psychological crisis levels.

\begin{table}[t]
\centering
\caption{Ablation results. Compare the full model with variants removing data augmentation, paralinguistic injection, auxiliary loss, and alternative paralinguistic modeling strategy (Emotion2vec Adaption).}
\label{tab:aux_loss}
\resizebox{1\linewidth}{!}{%
\begin{tabular}{lcc}
\toprule
\textbf{Configuration} & \textbf{Accuracy} & \textbf{Macro F1} \\
\midrule
Final Result    & \textbf{0.805} $\pm$ \textbf{0.061} & \textbf{0.802} $\pm$ \textbf{0.062} \\
w/o Augmentation   & 0.711 $\pm$ 0.115 & 0.702 $\pm$ 0.105 \\
w/o Paralinguistic Injection     & 0.764 $\pm$ 0.089 & 0.761 $\pm$ 0.088 \\
Emotion2vec Adaption & 0.772 $\pm$ 0.049 &  0.760 $\pm$ 0.065 \\
w/o Auxiliary Loss     & 0.790 $\pm$ 0.075 & 0.785 $\pm$ 0.075 \\

\bottomrule 
\end{tabular}
}
\end{table}

\section{Conclusion}
\label{sec:conclusion}

This paper proposes a novel framework for psychological crisis classification by fine-tuning LLM on authentic hotline calls. We introduced paralinguistic injection, which effectively capitalises on the superior text-processing capabilities of LLMs by mapping vital acoustic cues into textual markers, allowing LLM reasoning to jointly leverage semantics and affect,  achieving better performance than direct speech modelling. We further improved training via an auxiliary reasoning loss and data augmentation. Under 5-fold cross-validation, our best model achieved a robust macro F1-score of 0.802 (accuracy 0.805) for three-way classification, with ablation studies confirming the critical impact of both the paralinguistic features and the auxiliary reasoning loss.

\section{Acknowledgments}

We would like to thank the psychological support hotline center team led by Yang Luo and Yongsheng Tong for their expert annotation support and domain knowledge throughout this work. We are also grateful to Zengrui Jin for valuable guidance and insightful discussions during the project. We further acknowledge the computational support provided by the Multimedia Signal and Intelligent Information Processing Lab, Department of Electronic Engineering, Tsinghua University.

\section{Generative AI Use Disclosure}

Generative AI tools were used to assist with language editing, polishing, literature-related background research, and coding. All authors reviewed the final manuscript and are fully responsible for the content of the paper.

\bibliographystyle{IEEEtran}
\bibliography{mybib}

@book{CrisisStratChap1,
  author    = {Richard, K J and Julia, W and Rick, A M},
  title     = {Crisis Intervention Strategies, Chapter 1},
  edition   = {9},
  address   = {OH, USA},
  publisher = {Cengage},
  year      = {2025},
  pages     = {9--15}
}

@book{CrisisStratChap3,
  author    = {Richard, K J and Julia, W and Rick, A M},
  title     = {Crisis Intervention Strategies, Chapter 3},
  edition   = {9},
  address   = {OH, USA},
  publisher = {Cengage},
  year      = {2025},
  pages     = {55--84}
}

@article{su2025acoustic,
  title   = {Acoustic features for identifying suicide risk in crisis hotline callers: {M}achine learning approach},
  author  = {Su, Zhengyuan and Jiang, Huadong and Yang, Ying and Hou, Xiangqing and Su, Yanli and Yang, Li},
  journal = {Journal of Medical Internet Research},
  volume  = {27},
  pages   = {e67772},
  year    = {2025}
}

@article{broadbent2023machine,
  title   = {A machine learning approach to identifying suicide risk among text-based crisis counseling encounters},
  author  = {Broadbent, Meghan and Medina Grespan, Mattia and Axford, Katherine and Zhang, Xinyao and Srikumar, Vivek and Kious, Brent and Imel, Zac},
  journal = {Frontiers in psychiatry},
  volume  = {14},
  pages   = {1110527},
  year    = {2023}
}

@article{imel2024machine,
  title   = {Machine Learning--Based Evaluation of Suicide Risk Assessment in Crisis Counseling Calls},
  author  = {Imel, Zac E and Pace, Brian and Pendergraft, Brad and Pruett, Jordan and Tanana, Michael and Soma, Christina S and Comtois, Kate A and Atkins, David C},
  journal = {Psychiatric services},
  volume  = {75},
  number  = {11},
  pages   = {1068--1074},
  year    = {2024}
}

@inproceedings{feng2023large,
  title     = {Large Language Models Improve Alzheimer's Disease Diagnosis Using Multi-Modality Data},
  author    = {Feng, Yingjie and Xu, Xiaoyin and Zhuang, Yueting and Zhang, Min},
  booktitle = {Proc. {MedAI}},
  year      = {2023},
  address   = {Shanghai}
}

@inproceedings{ikeuchi2025efficient,
  title     = {Efficient and Effective Fine-tuning Method for Depression Detection from Conversation},
  author    = {Ikeuchi, Kenyu and Kishimoto, Taishiro and Nakai, Fumiya and Horigome, Toshiro and Kitazawa, Momoko and Ohtsuki, Tomoaki},
  booktitle = {Proc. {EMBC}},
  year      = {2025},
  address   = {Copenhagen}
}

@article{qorich2024advanced,
  title   = {Advanced deep learning and large language models for suicide ideation detection on social media},
  author  = {Qorich, Mohammed and El Ouazzani, Rajae},
  journal = {Progress in Artificial Intelligence},
  volume  = {13},
  number  = {2},
  pages   = {135--147},
  year    = {2024}
}

@article{deng2025evaluating,
  title   = {Evaluating {Large Language Models} in Crisis Detection: {A} Real-World Benchmark from Psychological Support Hotlines},
  author  = {Deng, Guifeng and Rao, Shuyin and Lin, Tianyu and Dai, Anlu and Wang, Pan and Xie, Junyi and Song, Haidong and Zhao, Ke and Xu, Dongwu and Cheng, Zhengdong and others},
  journal = {arXiv preprint arXiv:2506.01329},
  year    = {2025}
}

@inproceedings{teng2025enhancing,
  title     = {Enhancing depression detection with chain-of-thought prompting: {F}rom emotion to reasoning using large language models},
  author    = {Teng, Shiyu and Liu, Jiaqing and Jain, Rahul Kumar and Chai, Shurong and Hou, Ruibo and Tateyama, Tomoko and Lin, Lanfen and Chen, Yen-Wei},
  booktitle = {Proc. {EMBC}},
  year      = {2025},
  address   = {Copenhagen}
}

@article{patil2025cognitive,
  title   = {Cognitive-mental-{LLM}: {E}valuating reasoning in large language models for mental health prediction via online text},
  author  = {Patil, Avinash and Gedhu, Amardeep Kour},
  journal = {arXiv preprint arXiv:2503.10095},
  year    = {2025}
}

@article{zabelski2023crisis,
  title   = {Crisis lines: {C}urrent status and recommendations for research and policy},
  author  = {Zabelski, Sasha and Kaniuka, Andr{\'e}a R and A. Robertson, Ryan and Cramer, Robert J},
  journal = {Psychiatric services},
  volume  = {74},
  number  = {5},
  pages   = {505--512},
  year    = {2023}
}

@article{tyson2016preventing,
  title   = {Preventing Suicide and Self-Harm},
  author  = {Tyson, Philip and Law, Claire and Reed, Sophie and Johnsey, Emma and Aruna, Olusola and Hall, Sue},
  journal = {Crisis},
  volume  = {37},
  number  = {5},
  pages   = {353--360},
  year    = {2016}
}

@article{wang2020hotline,
  title   = {Hotline services in {China} during {COVID-19} pandemic},
  author  = {Wang, Jiali and Wei, Hualin and Zhou, Liang},
  journal = {Journal of affective disorders},
  volume  = {275},
  pages   = {125},
  year    = {2020}
}

@article{you2021study,
  title   = {A study on the competence characteristics of psychological hotline counselors during the outbreak of {COVID-19}},
  author  = {You, Linyu and Jia, Xiaoming and Ding, Yaping and An, Qin and Li, Bo},
  journal = {Frontiers in Psychology},
  volume  = {12},
  pages   = {566460},
  year    = {2021}
}

@inproceedings{eyben2010opensmile,
  title     = {{OpenSMILE}: {T}he {M}unich versatile and fast open-source audio feature extractor},
  author    = {Eyben, Florian and W{\"o}llmer, Martin and Schuller, Bj{\"o}rn},
  booktitle = {Proc. {ACM Multimedia}},
  year      = {2010},
  address   = {Firenze}
}

@article{hoffberg2020effectiveness,
  title     = {The effectiveness of crisis line services: a systematic review},
  author    = {Hoffberg, Adam S and Stearns-Yoder, Kelly A and Brenner, Lisa A},
  journal   = {Frontiers in public health},
  volume    = {7},
  pages     = {399},
  year      = {2020},
  publisher = {Frontiers Media SA}
}

@article{gao2022paraformer,
  title   = {Paraformer: {Fast} and {Accurate} {Parallel} {Transformer} for {Non}-autoregressive {End}-to-{End} {Speech} {Recognition}},
  author  = {Gao, Zhifu and Zhang, Shiliang and McLoughlin, Ian and Yan, Zhijie},
  journal = {arXiv preprint arXiv:2206.08317},
  year    = {2022}
}

@article{hu2022lora,
  title   = {{LoRA}: {Low}-{Rank} {Adaptation} of {Large} {Language} {Models}},
  author  = {Shen, Yelong and Wallis, Phillip and Allen-Zhu, Zeyuan and Li, Yuanzhi and Wang, Shean and others},
  journal = {arXiv preprint arXiv:2106.09685},
  year    = {2021}
}

@article{yang2024qwen2,
  title={Qwen2. 5 Technical Report},
  author={Yang, An and Yang, Baosong and Zhang, Beichen and Hui, Binyuan and Zheng, Bo and Yu, Bowen and Li, Chengyuan and Liu, Dayiheng and Huang, Fei and Wei, Haoran and others},
  journal={arXiv e-prints},
  pages={arXiv--2412},
  year={2024}
}

@article{xu2025qwen2,
  title={Qwen2. 5-{O}mni Technical Report},
  author={Xu, Jin and Guo, Zhifang and He, Jinzheng and Hu, Hangrui and He, Ting and Bai, Shuai and Chen, Keqin and Wang, Jialin and Fan, Yang and Dang, Kai and others},
  journal={arXiv preprint arXiv:2503.20215},
  year={2025}
}

@article{tian2025step,
  title={{Step-Audio-R1} Technical Report},
  author={Tian, Fei and Zhang, Xiangyu Tony and Zhang, Yuxin and Zhang, Haoyang and Li, Yuxin and Liu, Daijiao and Deng, Yayue and Wu, Donghang and Chen, Jun and Zhao, Liang and others},
  journal={arXiv preprint arXiv:2511.15848},
  year={2025}
}

@article{agarwal2025gpt,
  title={gpt-oss-120b \& gpt-oss-20b model card},
  author={Agarwal, Sandhini and Ahmad, Lama and Ai, Jason and Altman, Sam and Applebaum, Andy and Arbus, Edwin and Arora, Rahul K and Bai, Yu and Baker, Bowen and Bao, Haiming and others},
  journal={arXiv preprint arXiv:2508.10925},
  year={2025}
}

@article{eyben2015geneva,
  title={The {G}eneva minimalistic acoustic parameter set ({GeMAPS}) for voice research and affective computing},
  author={Eyben, Florian and Scherer, Klaus R and Schuller, Bj{\"o}rn W and Sundberg, Johan and Andr{\'e}, Elisabeth and Busso, Carlos and Devillers, Laurence Y and Epps, Julien and Laukka, Petri and Narayanan, Shrikanth S and others},
  journal={IEEE transactions on affective computing},
  volume={7},
  number={2},
  pages={190--202},
  year={2015},
}

@inproceedings{wang2025benchmarking,
  title={Benchmarking Contextual and Paralinguistic Reasoning in Speech-{LLM}s: {A} Case Study with In-the-Wild Data},
  author={Wang, Qiongqiong and Sailor, Hardik Bhupendra and Liu, Tianchi and Zhang, Wenyu and Huzaifah, Muhammad and Lertcheva, Nattadaporn and Sun, Shuo and Chen, Nancy F and Wu, Jinyang and Aw, AiTi},
  booktitle={Proc. {EMNLP}},
  year={2025},
  address={Suzhou}
}

@inproceedings{wang2025incorporating,
  title={Incorporating contextual paralinguistic understanding in large speech-language models},
  author={Wang, Qiongqiong and Sailor, Hardik B and Wong, Jeremy HM and Liu, Tianchi and Sun, Shuo and Zhang, Wenyu and Huzaifah, Muhammad and Chen, Nancy and Aw, Ai Ti},
  booktitle={Proc. ASRU},
  year={2025},
  address = {Honolulu}
}

@inproceedings{ma2024emotion2vec,
  title={emotion2vec: {S}elf-supervised pre-training for speech emotion representation},
  author={Ma, Ziyang and Zheng, Zhisheng and Ye, Jiaxin and Li, Jinchao and Gao, Zhifu and Zhang, Shiliang and Chen, Xie},
  booktitle={Proc. {ACL}},
  year={2024},
  address = {Bangkok}
  
}

@article{chen2024deep,
  title={Deep learning and large language models for audio and text analysis in predicting suicidal acts in Chinese psychological support hotlines},
  author={Chen, Yining and Li, Jianqiang and Song, Changwei and Zhao, Qing and Tong, Yongsheng and Fu, Guanghui},
  journal={arXiv preprint arXiv:2409.06164},
  year={2024}
}

@article{song2024exploratory,
  title={An exploratory deep learning approach for predicting subsequent suicidal acts in chinese psychological support hotlines},
  author={Song, Changwei and Zhao, Qing and Li, Jianqiang and Chen, Yining and Tong, Yongsheng and Fu, Guanghui},
  journal={arXiv preprint arXiv:2408.16463},
  year={2024}
}

@inproceedings{panayotov2015librispeech,
  title={Librispeech: an asr corpus based on public domain audio books},
  author={Panayotov, Vassil and Chen, Guoguo and Povey, Daniel and Khudanpur, Sanjeev},
  booktitle={Proc. {ICASSP}},
  year={2015},
  address = {Brisbane}
}

\end{document}